\documentclass[letterpaper]{article} 
\usepackage{aaai2026}  
\usepackage{times}  
\usepackage{helvet}  
\usepackage{courier}  
\usepackage[hyphens]{url}  
\usepackage{graphicx} 
\usepackage{natbib}  
\usepackage{caption} 
\usepackage{algorithm}
\usepackage{algorithmic}

\usepackage{amsmath}
\usepackage{booktabs}
\usepackage{arydshln}
\usepackage{newfloat}
\usepackage{listings}
\DeclareCaptionStyle{ruled}{labelfont=normalfont,labelsep=colon,strut=off} 
\floatstyle{ruled}
\newfloat{listing}{tb}{lst}{}
\floatname{listing}{Listing}
\title{Knowledge Boundary Discovery for Large Language Models}

\author{
    Ziquan Wang\textsuperscript{\rm 1,\rm 2},
    Zhongqi Lu\textsuperscript{\rm 1,\rm 2}\thanks{Corresponding author.}
}
\affiliations{
    \textsuperscript{\rm 1}College of Artificial Intelligence, China University of Petroleum-Beijing, China\\
    \textsuperscript{\rm 2}Hainan Institute of China University of Petroleum (Beijing), Sanya, Hainan, China\\
    2021011537@student.cup.edu.cn,
    zhongqi@cup.edu.cn
}

\begin{document}

\maketitle

\begin{abstract}
    We propose \textbf{Knowledge Boundary Discovery (KBD)}, a reinforcement learning based framework to explore the knowledge boundaries of the Large Language Models (LLMs). We define the knowledge boundary by automatically generating two types of questions: (i) those the LLM can confidently answer (\textit{within-knowledge boundary}) and (ii) those it cannot (\textit{beyond-knowledge boundary}). Iteratively exploring and exploiting the LLM's responses to find its knowledge boundaries is challenging  because of the hallucination phenomenon. To find the knowledge boundaries of an LLM, the agent interacts with the LLM under the modeling of exploring a partially observable environment. The agent generates a progressive question as the action, adopts an entropy reduction as the reward, receives the LLM's response as the observation and updates its belief states. We demonstrate that the KBD detects knowledge boundaries of LLMs by automatically finding a set of non-trivial answerable and unanswerable questions. We validate the KBD by comparing its generated knowledge boundaries with manually crafted LLM benchmark datasets. Experiments show that our KBD-generated question set is comparable to the human-generated datasets. Our approach paves a new way to evaluate LLMs.
\end{abstract}


\section{Introduction}

Large Language Models (LLMs) have made remarkable strides across diverse domains \cite{NEURIPS2020_1457c0d6}. Despite their impressive capabilities, a persistent challenge remains: LLMs often produce vague, inaccurate, or incorrect responses, commonly referred to as ``hallucinations". Hallucinations arise primarily due to three reasons: (i) the LLM possesses relevant knowledge but generates unqualified responses, (ii) the LLM lacks the necessary knowledge but attempts to answer, and (iii) the LLM has no required knowledge. A key underlying issue is the LLM’s inability to recognize its own \textit{knowledge boundaries}, leading to overconfidence and unreliable answers beyond its expertise.

Prompt engineering is one major method to address hallucinations. While being effective when the LLM has sufficient knowledge, the prompt engineering approach struggles when the LLM does not have any related knowledge. In other words, prompt engineering can only handle the cases when the LLM possesses relevant knowledge but generates unqualified responses. The measure of LLMs' knowledge level is essential to the proper post processing. Therefore, various manually crafted benchmarks are published. However, even though many human efforts have been involved in generating these datasets, the static benchmarks fail to adapt dynamically to the LLM’s evolving responses. To fully understand the knowledge capabilities of LLMs, it is necessary to engage in iterative interactions with the LLM, systematically probing and exploring its knowledge boundaries. This aligns naturally with reinforcement learning (RL) methods.

In this work, we define the knowledge boundary of LLMs by automatically generating two types of questions: those LLM can confidently and accurately answer (\textit{within-knowledge boundary}) and those it cannot (\textit{beyond-knowledge boundary}). This definition provides a more intuitive way to gauge the extent of an LLM’s knowledge.

We propose \textbf{Knowledge Boundary Discovery (KBD)}, a novel reinforcement learning-based framework designed to detect LLMs' knowledge boundaries. Due to the \textit{partially observable} nature of the environment, the agent must infer the hidden belief of the LLM’s knowledge level based on the observation of its responses. Therefore, we adopt the Partially Observable Markov Decision Process (POMDP) modeling in our KBD framework. The KBD employs a reinforcement learning agent to iteratively generate progressive questions around the knowledge boundaries of the LLM, adopt an entropy reduction as the reward, receive the LLM’s response as the observation and update the belief states.

To guide the search for the knowledge boundaries, we integrate \textit{information gain} from information theory into the reward function. Information gain quantifies the reduction in uncertainty achieved through each interaction, aligning the rewards with the quality of the LLM’s responses. Our intuition is that the information gain changes dramatically around the knowledge boundaries. This reward function allows the agent to identify areas where the LLM demonstrates stronger knowledge, as evidenced by decreasing the response entropy. Through iterative interactions, KBD dynamically detects the knowledge boundaries.

Moreover, our KBD-generated questions are \textit{non-trivial}—they are not random or template-based, but are semantically meaningful and reside near the boundary where the LLM’s knowledge becomes uncertain. This property is empirically supported by entropy-based clustering and embedding-space distance analysis, which shows that KBD-generated questions differ significantly from random ones while staying close to the knowledge boundary between known and unknown questions.

In summary, our contributions are as follows:
\begin{itemize}
\item We introduce a practical and intuitive definition of the knowledge boundary for LLMs, covering both confidently answerable and unanswerable questions, thereby offering a clear and operational criterion for evaluating LLM capabilities.
\item We propose \textbf{Knowledge Boundary Discovery (KBD)}, a novel reinforcement learning framework that dynamically identifies knowledge boundaries via entropy-guided exploration. Experimental results demonstrate that KBD effectively discovers knowledge boundaries and generates question sets that are comparable in quality to human-curated benchmarks.
\end{itemize}

\section{Related Work}

\subsection{Knowledge Boundaries and Hallucinations in LLMs}

Understanding the knowledge boundaries of Large Language Models (LLMs) is essential for recognizing their limitations and guiding their safe and effective deployment. These boundaries encompass prompt-agnostic, prompt-sensitive, and unanswerable knowledge types \cite{zhang2024defining,yin2024benchmarking,li2024knowledgeboundarylargelanguage}. Despite their fluency and impressive performance, LLMs frequently suffer from hallucinations, i.e., producing confident but factually incorrect responses, stemming from limitations in training data, model architecture, and inference mechanisms \cite{maynez2020,ye2023,zhang2023}. Such issues are further exacerbated by outdated or noisy data used during training \cite{penedo2023,reddy2024}.

To characterize the knowledge limitations of LLMs, recent studies have investigated how LLMs handle known versus unknown information, including dynamically shifting knowledge states. These works show that LLMs often overestimate their abilities, failing to recognize when a question is unanswerable \cite{yin2023largelanguagemodelsknow,amayuelas2024knowledgeknowledgeexploringknownunknowns,liu2024examiningllmsuncertaintyexpression}. Additional research has analyzed the effect of retrieval augmentation, showing that the quality of retrieved information directly influences the accuracy and reliability of LLM outputs \cite{zhang2024,yin2023largelanguagemodelsknow,ren2024investigatingfactualknowledgeboundary}.

To detect and mitigate hallucinations, several approaches have been developed to detect when an LLM generates hallucinated content. Pacchiardi~\shortcite{pacchiardi2023} proposed a simple lie detector to detect inaccurate answers. Chen~\shortcite{chen2024inside} introduced the EigenScore metric to assess the internal consistency of responses, while Zhang~\shortcite{zhang2024selfalignmentfactualitymitigatinghallucinations} developed SELF-EVAL, a mechanism that enables LLMs to self-verify their factual output using internal knowledge alone.

To explicitly model and teach knowledge boundaries, other works aim to improve how LLMs recognize and express the limits of their knowledge. Chen~\shortcite{chen2024teachinglargelanguagemodels} proposed COKE, a method that leverages internal confidence signals to teach LLMs to better recognize their knowledge boundaries. Pezeshkpour~\shortcite{pezeshkpour2023measuringmodifyingfactualknowledge} employed information-theoretic measures like entropy and KL-divergence to more accurately model factual confidence, helping detect hallucinations and improve retrieval-augmented generation. Zheng~\shortcite{zheng2024kglensefficienteffectiveknowledge} introduced KGLens, a sampling-based framework that uses knowledge graphs and Thompson sampling to efficiently identify factual blind spots in LLMs.

\subsection{Reinforcement Learning for Knowledge Exploration}

Reinforcement learning (RL) has advanced knowledge exploration by integrating querying mechanisms, goal conditioning, and large language models (LLMs). Approaches like Asking for Knowledge (AFK) enable RL agents to efficiently query external knowledge sources, addressing sparse rewards and large action spaces \cite{liu2022askingknowledgetrainingrl}. Goal-conditioned methods, such as ReenGAGE, use knowledge distillation to transfer information and prioritize goals in high-dimensional spaces \cite{levine2023goalconditionedqlearningknowledgedistillation}. Model-free techniques, such as RandQL, employ randomized learning rates for efficient exploration without Bayesian complexity \cite{tiapkin2023modelfreeposteriorsamplinglearning}. LLMs are increasingly integrated into RL workflows, as seen in TWOSOME and other frameworks that align language models with decision-making tasks through fine-tuning and structured prompting \cite{tan2024trueknowledgecomespractice,gholamian2024reinforcementlearningproblemsolving}. Extreme Q-Learning (X-QL) further enhances RL by applying Extreme Value Theory to model Q-values and mitigate function approximation errors \cite{garg2023extremeqlearningmaxentrl}. These innovations enable autonomous agents to query, reason, and solve complex, knowledge-intensive problems.

\subsection{Retrieval-Augmented Generation}
Retrieval-Augmented Generation (RAG) enhances generative models by integrating external knowledge retrieval. Relevant to our work are contributions to data augmentation and selection strategies.
Lewis~\shortcite{lewis2021retrievalaugmentedgenerationknowledgeintensivenlp} proposed the RAG framework, using the Dense Passage Retriever (DPR) to dynamically retrieve Top-k documents for improved contextual coverage. Zhang~\shortcite{zhang-etal-2024-retrievalqa} introduced RetrievalQA, which adaptively identifies whether retrieval is necessary, reducing computational overhead. Yue~\shortcite{yue2024retrievalaugmentedfactverification} designed a framework that retrieves evidence to synthesize contrastive arguments, supporting or refuting claims.
Yu~\shortcite{yu-etal-2023-retrieval} addressed few-shot learning with retrieval objectives that prioritize task-relevant examples. Chen~\shortcite{chen2023decouplingknowledgememorizationretrievalaugmented} proposed RETROPROMPT, which retrieves training examples from an internal knowledge store to reduce reliance on parametric memory. Su~\shortcite{su2024dragindynamicretrievalaugmented} propose DRAGIN, a dynamic RAG framework that can dynamically determine when to trigger the retrieval.

In conclusion, our KBD method, combining reinforcement learning with entropy-based measures, demonstrates significant advantages in dynamic question generation and automated validation. It not only enhances the ability to explore knowledge boundaries, but also effectively reduces response uncertainty.

\section{Methodology}

\subsection{Reinforcement Learning Framework for KBD}

KBD formulates the discovery of LLM knowledge boundaries as a reinforcement learning (RL) task. In this setup, an agent interacts with the target LLM, treated as a black-box environment, by generating questions (actions) and receiving observations. Each response is evaluated through entropy-based reward signals, guiding the agent toward areas of high or low uncertainty in model confidence.

Since the internal state of an LLM is inaccessible, the agent maintains a belief state based on observed responses. This naturally aligns the problem with a Partially Observable Markov Decision Process (POMDP), enabling a structured exploration of the LLM’s latent knowledge.

\subsection{POMDP Formulation}

We formalize the task of exploring LLM knowledge boundaries as a Partially Observable Markov Decision Process (POMDP), represented by the tuple \( (\mathcal{S}, \mathcal{A}, \mathcal{T}, \mathcal{R}, \Omega, \mathcal{O}) \):

\begin{itemize}
    \item \(\mathcal{S}\): The set of states representing the knowledge level of the target LLM. Note that the true state \(s \in \mathcal{S}\) is hidden and cannot be directly observed (e.g., whether the question within-knowledge boundary or beyond-knowledge boundary). Instead, we use the belief state \(b\) as an estimate of \(s\) for the computation. \(b\) is a probability distribution over \(\mathcal{S}\), representing the probability distribution of the true states for each historical question, which is iteratively updated based on actions and observations.
    \item \(\mathcal{A}\): The set of possible actions(generated questions).
    \item \(\mathcal{T}(b'|b, a)\): The state transition function.
    \item \(\mathcal{R}(b, a)\): The reward function(entropy-based).
    \item \({\Omega}\): The set of all possible observations corresponding to the responses provided by the target LLM.
    \item \(\mathcal{O}(o|b, a)\): The observation function.
\end{itemize}

This framework captures two essential aspects of interacting with the target LLMs:

\begin{itemize}
    \item \textbf{Partial Observability}: The agent cannot access the internal computations or knowledge representation of the target LLM and must infer its state based on observed responses.
    \item \textbf{Sequential Decision-Making}: Each action (question) influences future observations and rewards, necessitating a strategy to balance exploration (investigating new areas) and exploitation (focusing on areas where the target LLM performs well).
\end{itemize}

\paragraph{Belief State Encoder}

To model the belief state \( b_t \), we define an encoder function \( f(\cdot) \) that computes a probability distribution over the hidden true state \(s\) (i.e., whether each question belongs to the within-knowledge boundary or the beyond-knowledge boundary) based on the entropy of the responses. For each interaction pair \((a_i, o_i)\), \( E_i \) denote the response entropy of \( o_i \). The probability that this interaction corresponds to a within-knowledge boundary is defined as:

\begin{equation}
P(s_i = \texttt{within} \mid E_i) = \sigma(\beta (E_{\text{th}} - E_i)),
\label{eq:belief_estimation}
\end{equation}

where:
\begin{itemize}
    \item \( \sigma(\cdot) \) is the sigmoid function.
    \item \( E_{\text{th}} \) is the entropy threshold that separates within and beyond boundaries (e.g., 40 and 170 in our experiments).
    \item \( \beta \) is a scaling hyperparameter that controls the sharpness of the boundary.
\end{itemize}

The belief state at time \( t \), denoted as \( b_t \), is then a collection of these probabilities for each interaction up to time \( t \):

\begin{equation}
\begin{aligned}
b_t = \{ &P(s_1 = \texttt{within} \mid E_1), \dots, \\
         &P(s_{t-1} = \texttt{within} \mid E_{t-1}) \}.
\end{aligned}
\end{equation}

This probabilistic representation allows the agent to track and update its belief about the LLM’s knowledge boundary based on the entropy trends in the dialogue history.

\paragraph{Action Space}
The action space \(\mathcal{A}\) consists of all possible questions the agent can pose to the target LLM. At each time step \(t\), the agent maintains a belief state \(b_t\), and uses a learned policy \(\pi\bigl(a_t \mid b_t\bigr)\) to select an action \(a_t \in \mathcal{A}\). This policy ensures that the generated questions are contextually relevant and strategically guide the exploration of the target LLM’s knowledge boundaries, by aiming to refine the agent's belief about the LLM’s internal knowledge state.

\paragraph{Observation}

An observation \( o_t \) is the response provided by the target LLM when the agent’s question \( a_t \) is posed:

\begin{equation}
o_t = \text{LLM}_{\text{target}}(a_t | b_t).
\end{equation}

Observations include the textual content of the target LLM’s response and its log-probability (logprob) in the vocabulary. This logprob is crucial for calculating the response entropy, which serves as a measure of the uncertainty of the target LLM.
\paragraph{Relationship between Belief State, Action, and Observation}

The belief state \( b_t \) summarizes the agent’s understanding of the target LLM’s knowledge level up to time \( t \). After the agent takes action \( a_t \) and observes the response \( o_t \), the belief state transitions to \( b_{t+1} \), incorporating the new information:
\begin{equation}
b_{t+1} \leftarrow b_t \cup \{ (a_t, o_t) \}.
\label{eq:state_transition}
\end{equation}

This iterative structure ensures that the agent’s decisions are informed by the cumulative interaction history, enabling adaptive refinement of its questioning strategy.

\paragraph{Entropy-Based Reward}
To encourage discovery of boundaries, we reward changes in entropy:

\begin{equation}
r_t = \left| E_{\text{prev}} - E_{\text{current}} \right|,
\label{eq:reward_function}
\end{equation}

\begin{itemize}
    \item \( E_{\text{current}} \) is the entropy of the target LLM's response at time \( t \).
    \item \( E_{\text{prev}} \) is the entropy of the target LLM's previous response at time \( t-1 \).
\end{itemize}

The entropy \( E \) uses the logprob over the target LLM’s vocabulary \( V \):

\begin{equation}
E = - \sum_{w \in V} P(w) \log P(w),
\label{eq:entropy}
\end{equation}

where:

\( P(w) \) is the probability assigned to word \( w \). By incentivizing entropy reduction, the agent is motivated to pose questions that clarify the target LLM’s responses and reduce uncertainty. Lower entropy indicates high LLM confidence (\textit{within}), while high entropy signals uncertainty (\textit{beyond}).


\subsection{Q-Learning with Belief States}

In a partially observable setting, the true state \(s\) of the environment is hidden and cannot be directly observed. Instead, the agent maintains a \emph{belief state} \(b\), which is an estimated probability distribution over all possible states \(s \in \mathcal{S}\). The belief state serves as an approximation of the true state, summarizing the agent's knowledge of the environment based on its actions and observations. We define the Q-function as:
\begin{equation}
  Q^*(b, a) 
  \;=\; 
  {E}\bigl[R(b, a) + \gamma \max_{a'} Q^*(b', a') \mid b\bigr],
  \label{eq:q_value_belief_standard}
\end{equation}
where \(\gamma \in [0,1)\) is the discount factor, and \({E}[\cdot \mid b]\) takes the expectation with respect to the belief $b\sim$ the hidden state \(s\).  At each time step \(t\):

\begin{enumerate}
    \item The agent has a current belief state \(b_t\).
    \item Choose an action \(a_t\) using \(\epsilon\)-greedy policy over \(Q(b_t,a)\).
    \item It executes \(a_t\) and receives observation \(o_t\) (the LLM’s response).
    \item Calculate the immediate reward \(r_t\) (e.g., the change in response entropy).
    \item Update its belief state to \(b_{t+1}\).
    \item Update the Q-value \(Q_t(b_t, a_t)\):
    \begin{equation}
    \begin{split}
    Q_{t+1}(b_t, a_t) \;\leftarrow\; Q_t(b_t, a_t) 
    \;+\; \alpha \Bigl[ r_t 
    \;+\; \\ \gamma \max_{a'} Q_t(b_{t+1}, a') 
    \;-\; Q_t(b_t, a_t) \Bigr],
    \end{split}
    \label{eq:q_update_standard}
    \end{equation}
    where \(\alpha\) is the learning rate.
\end{enumerate}

In this Q-learning framework, the agent not only benefits from the immediate reward \(r_t\), but also from the potential future rewards, captured by the term \(\gamma \max_{a'} Q_t(b_{t+1}, a')\).

\paragraph{\(\epsilon\)-Greedy Action Selection}

To balance exploration and exploitation, the agent adopts an \(\epsilon\)-greedy action selection strategy:
\begin{equation}
  a_t \;=\;
  \begin{cases}
    \text{random action from } \mathcal{A}, & \text{with probability } \epsilon, \\
    \displaystyle
    \arg\max_{a \in \mathcal{A}} Q(b_t, a), & \text{with probability } 1 - \epsilon,
  \end{cases}
  \label{eq:epsilon_greedy}
\end{equation}
where \( \mathcal{A} \) denotes the action space (i.e., candidate questions), and \( Q(b_t, a) \) is the estimated value of taking action \( a \) under belief state \( b_t \).

Exploration (\(\epsilon\) branch) injects stochastic questions that can push the dialogue into regions the agent has not yet sampled, which is crucial to uncovering \textit{ beyond-knowledge boundary} gaps.  
Exploitation (greedy branch) focuses on high-value questions expected to maximize entropy reduction, refining the \textit{within-knowledge boundary} area. Each episode begins with a randomly initialized question, which may fall either within or beyond the LLM’s knowledge boundary. Even if a single episode fails to reach the beyond-knowledge region, the aggregated effect across many episodes enables a comprehensive boundary exploration. 

Moreover, our method supports flexible reward shaping and policy guidance. By modifying the reward function (e.g., rewarding high-entropy responses) and adjusting the exploration bias, the agent can be directed to actively search for questions that fall into the \textit{beyond-knowledge boundary}. This design ensures that the agent systematically probes both sides of the knowledge boundary, reinforcing the dual discovery capability of our proposed algorithm.

\begin{algorithm}[t]
\caption{Updating Strategy for KBD}
\label{alg:q_learning_entropy_pomdp}
\begin{algorithmic}[1]
\REQUIRE Learning rate \(\alpha\), exploration rate \(\epsilon\), discount factor \(\gamma\), number of episodes \(N\), maximum steps per episode \(M\), entropy threshold \(E_{\text{threshold}}\) and topic \(T\)
\STATE \textbf{Initialize} \(Q(b, a) \leftarrow 0\) for all belief states \(b\) and actions \(a\)
\FOR{episode \( = 1 \) to \(N\)}
    \STATE \textbf{Initialize} prior belief \( b_1 \) (e.g., uniform or from domain knowledge) with topic \(T\)
    \STATE \textbf{Initialize} previous entropy \( E_{\text{prev}} \leftarrow \infty \)
    \FOR{\( t = 1 \) to \(M\)}
        \STATE With probability \(\epsilon\), select a random action \( a_t \in \mathcal{A} \)
        \STATE Otherwise, select \( a_t \leftarrow \arg\max_{a} Q(b_t, a) \)
        \STATE \textbf{Execute} \( a_t \), \textbf{receive observation} \( o_t \) from the LLM
        \STATE \textbf{Compute current entropy} \( E_{\text{current}} \) from \( o_t \)
        \STATE \textbf{Calculate reward}: \(r_t = |E_{\text{prev}} - E_{\text{current}}| \)
        \STATE \textbf{Update Q-value} by Eq.~\ref{eq:q_update_standard}
        
        \STATE \textbf{Update belief state} \( b_{t+1} \leftarrow b_t \cup \{ (b_t, o_t) \} \)
        \STATE Update previous entropy: \( E_{\text{prev}} \leftarrow E_{\text{current}} \)
        \IF{ \( E_{\text{current}} < E_{\text{threshold}} \) }
            \STATE \textbf{break}
        \ENDIF
    \ENDFOR
\ENDFOR
\end{algorithmic}
\end{algorithm}

\section{Experiments}

In this section, we first validate the effectiveness of using entropy to define and identify the knowledge boundaries of LLMs, and demonstrate that the discovered boundaries are meaningful and non-trivial. We then show that the question set generated by our KBD method achieves results comparable to human-generated datasets used in previous work.

\begin{table*}[t]
\renewcommand{\arraystretch}{1.2}
\setlength{\tabcolsep}{5pt}
\centering
\begin{tabular}{lccc|ccc|ccc}
\toprule
\textbf{Method} 
& \multicolumn{3}{c|}{\textbf{KUQ}} 
& \multicolumn{3}{c|}{\textbf{Sware}} 
& \multicolumn{3}{c}{\textbf{Infeasible Benchmark}} \\

& $\mathbf{K_{aware}}$ & $\mathbf{U_{aware}}$ & $\mathbf{S_{aware}}$ 
& $\mathbf{K_{aware}}$ & $\mathbf{U_{aware}}$ & $\mathbf{S_{aware}}$
& $\mathbf{K_{aware}}$ & $\mathbf{U_{aware}}$ & $\mathbf{S_{aware}}$ \\ 
\midrule
Min-Prob     & 78.8 & 24.0 & 51.4 & 82.5 & 17.3 & 49.9 & 76.3 & 25.6 & 51.0 \\
Fst-Prob     & 51.0 & 36.9 & 43.9 & 94.4 &  5.2 & 49.8 & 65.5 & 27.1 & 41.3 \\
Prod-Prob    & 48.6 & 61.0 & 54.8 & 95.4 & 30.6 & 63.0 & 86.4 & 18.7 & 52.5 \\
Prior        & 83.0 & 40.1 & 61.5 & 80.8 & 37.7 & 59.2 & 74.8 & 42.9 & 58.9 \\
Posterior    & 76.0 & 17.3 & 46.6 & 83.0 & 16.1 & 49.5 & 85.2 & 17.4 & 51.3 \\
IC-IDK       & 88.0 & 13.5 & 50.7 & 80.5 & 19.2 & 49.8 & 90.9 & 18.5 & 54.7 \\
Verb         & 95.5 & 15.9 & 55.7 & 99.0 & 20.8 & 59.9 & 96.1 & 30.2 & 63.1 \\
Entropy      & 61.4 & 78.2 & \textbf{69.8} 
             & 76.7 & 56.5 & \textbf{66.6} 
             & 80.1 & 60.3   & \textbf{70.2} \\ 
\bottomrule
\end{tabular}
\caption{Performance comparison of entropy confidence estimation with several baselines on three human-generated datasets: KUQ, Sware and Infeasible Benchmark. The metrics include: \(K_{\text{aware}}\) (accuracy on answerable questions), \(U_{\text{aware}}\) (rate of correctly indicating "unknown" for unanswerable questions), and \(S_{\text{aware}} = \tfrac{1}{2}(K_{\text{aware}} + U_{\text{aware}})\), which reflects overall self-awareness. Entropy consistently achieves the highest \(S_{\text{aware}}\) on all datasets, and shows notably strong performance in \(U_{\text{aware}}\), demonstrating its effectiveness at identifying unanswerable questions and delineating knowledge boundaries.}
\label{table:confidence_method_comparison}
\end{table*}

\subsection{Experimental Setup}

To evaluate the effectiveness of our proposed KBD framework, we adopt a dual-model architecture comprising: (1) a powerful supervised model responsible for question generation and (2) smaller target LLMs whose knowledge boundaries are to be explored. Importantly, the supervised model only generates questions and does not participate in answering them, ensuring that the identified knowledge boundaries reflect the limitations of the target models alone.

\begin{itemize}
    \item \textbf{Target LLMs:} We examine the knowledge boundaries of three target models: ChatGLM3-6B, ChatGLM2-6B (both from Zhipu AI~\cite{glm2024chatglmfamilylargelanguage}), and LLaMA 7B-Chat. These models serve as the environments that interact with the RL agent to reveal their knowledge boundaries.
    
    \item \textbf{Supervised Model:} We use the more powerful GLM-4-9b model~\cite{glm2024chatglmfamilylargelanguage} as the supervised model to generate questions. Due to its larger parameter scale and broader coverage, we assume that GLM-4-9B has a wider knowledge boundary than the target models. 
\end{itemize}

In each episode, the supervised model proposes a question to the target LLM, then the target LLM answers. And based on the entropy of the response, the agent updates its strategy. Through repeated interaction, the agent progressively identifies regions where the target model answers with high confidence (within-knowledge boundary) versus high uncertainty (beyond-knowledge boundary).

\begin{figure}[t]
    \centering
    \includegraphics[width=\linewidth]{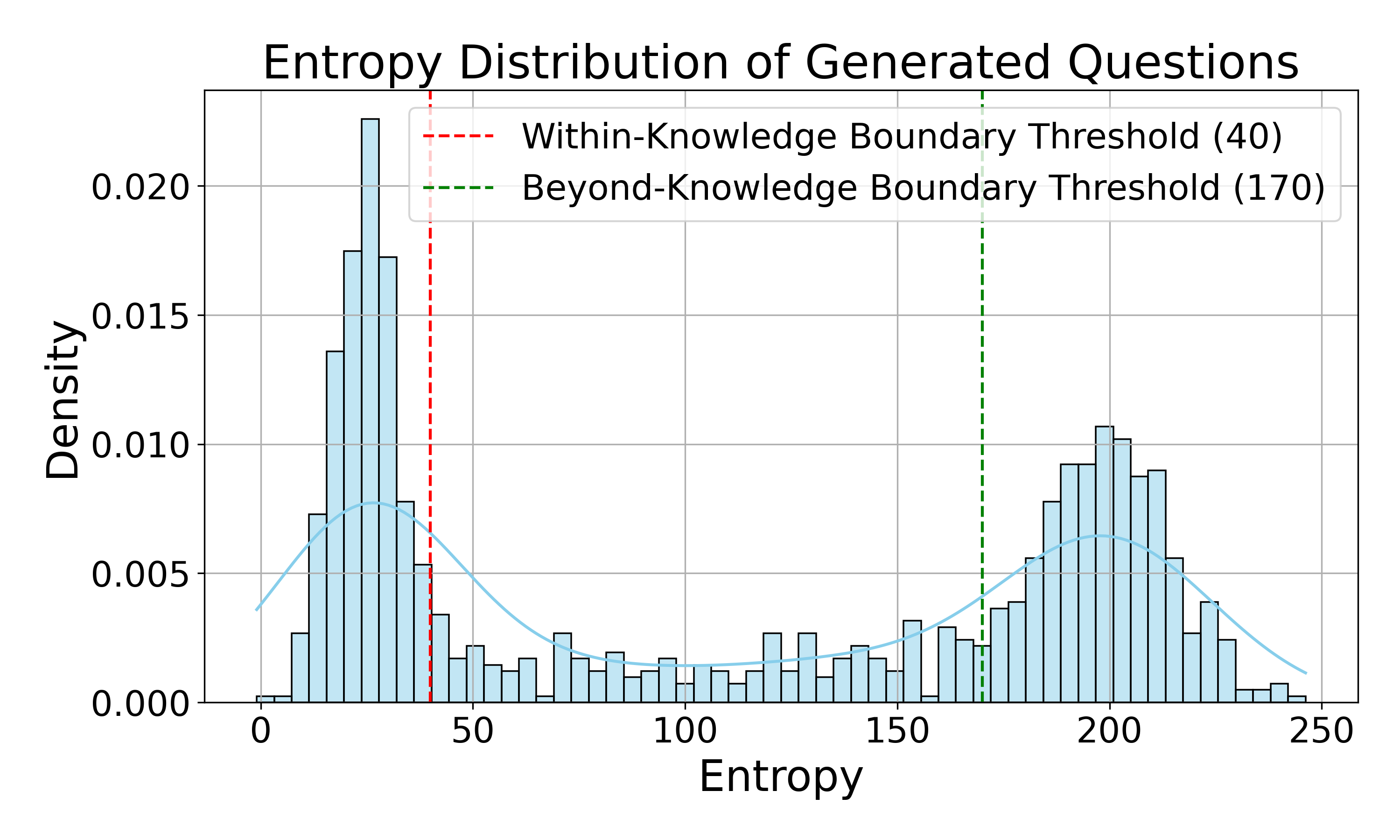}
    \caption{Distribution of entropy values for over 2,000 responses generated by our KBD algorithm across various topics and parameter configurations. The histogram and KDE curve reveal two prominent peaks: one in the low-entropy region (\(\leq 40\)), and another in the high-entropy region (\(\geq 170\)). These correspond to within-knowledge and beyond-knowledge boundaries, respectively. Only about 20\% of the questions fall into the mid-entropy range (\(40 < \text{entropy} < 170\)), suggesting that the transition zone (i.e. where the model’s knowledge is ambiguous) is narrow. }
    \label{fig:boundary}
\end{figure}

\begin{table*}[h!]
\centering
\begin{tabular}{p{8cm}p{8cm}}
\hline
\textbf{Within-Knowledge Boundary} & \textbf{Beyond-Knowledge Boundary} \\ \hline
Portico of the Aetolians and Delphi refer to what culture? & What challenges might future telemedicine face? \\ 
Entropy: 26.30 & Entropy: 205.71 \\ \hline

Which Enlightenment thinker was against the separation of powers? & Are there limits to human creativity? \\ 
Entropy: 27.84 & Entropy: 211.29 \\ \hline

What treaty ended the Russo-Persian War? & What sort of life exists in the center of the cosmos? \\ 
Entropy: 26.63 & Entropy: 197.64 \\ \hline
\end{tabular}
\caption{Random sample from the KBD-generated dataset. Questions with entropy values \(\leq 40\) are classified as within-knowledge boundary, while those with entropy values \(\geq 170\) are categorized as beyond-knowledge boundary.}
\label{table:KBD_question_set_structure}
\end{table*}

\subsection{Effectiveness of Entropy as a Confidence Estimation Metric}
\label{entropy_evaluation}

Confidence estimation serves as a critical tool for evaluating whether an LLM's response falls within or beyond its knowledge boundaries. In our framework, entropy plays a central role as a unified metric that captures both the confidence of the model and the depth of its knowledge.

We first evaluate entropy as a confidence estimation metric by comparing it with several existing baselines using three human-generated datasets: KUQ~\cite{amayuelas2024knowledgeknowledgeexploringknownunknowns}, Sware~\cite{yin2023largelanguagemodelsknow} and Infeasible Benchmark~\cite{zhang2024defining}. Following the evaluation setup of Chen~\shortcite{chen2024teachinglargelanguagemodels}, we include the following baseline methods: Min-Prob, Prod-Prob, Fst-Prob, Prior Prompt~\cite{ren2024investigatingfactualknowledgeboundary}, Posterior Prompt~\cite{kadavath2022languagemodelsmostlyknow}, In-Context IDK (IC-IDK)~\cite{cohen-etal-2023-lm}, and Verbalized Uncertainty (Verb)~\cite{tian2024finetuning}.

As shown in Table~\ref{table:confidence_method_comparison}, entropy achieves competitive or superior performance on all metrics. In particular, it excels in identifying unanswerable questions (\(U_{\text{aware}}\)) and achieves the highest overall self-awareness score (\(S_{\text{aware}}\)) on both datasets. These results demonstrate the effectiveness of entropy in identifying unanswerable questions and delineating knowledge boundaries, outperforming probability-based confidence scores in capturing model uncertainty.

Furthermore, to illustrate how entropy delineates the knowledge boundary, we analyze the entropy distribution over a large set of KBD-generated questions. Figure~\ref{fig:boundary} overlays a histogram with a kernel density estimation (KDE) curve (i.e., a smooth nonparametric density estimate obtained by summing Gaussian kernels over all samples). The result is a clean bimodal distribution: one mode centered around entropy \( \leq 40 \) (answerable) and the other beyond \( \geq 170 \) (unanswerable), with a narrow transition region between. This empirical evidence reinforces our threshold selection and validates entropy’s effectiveness in marking clear knowledge boundaries. The detailed procedure for selecting the boundary thresholds based on entropy is provided in Appendix A.

\begin{figure}[t]
    \centering
    \includegraphics[width=\linewidth]{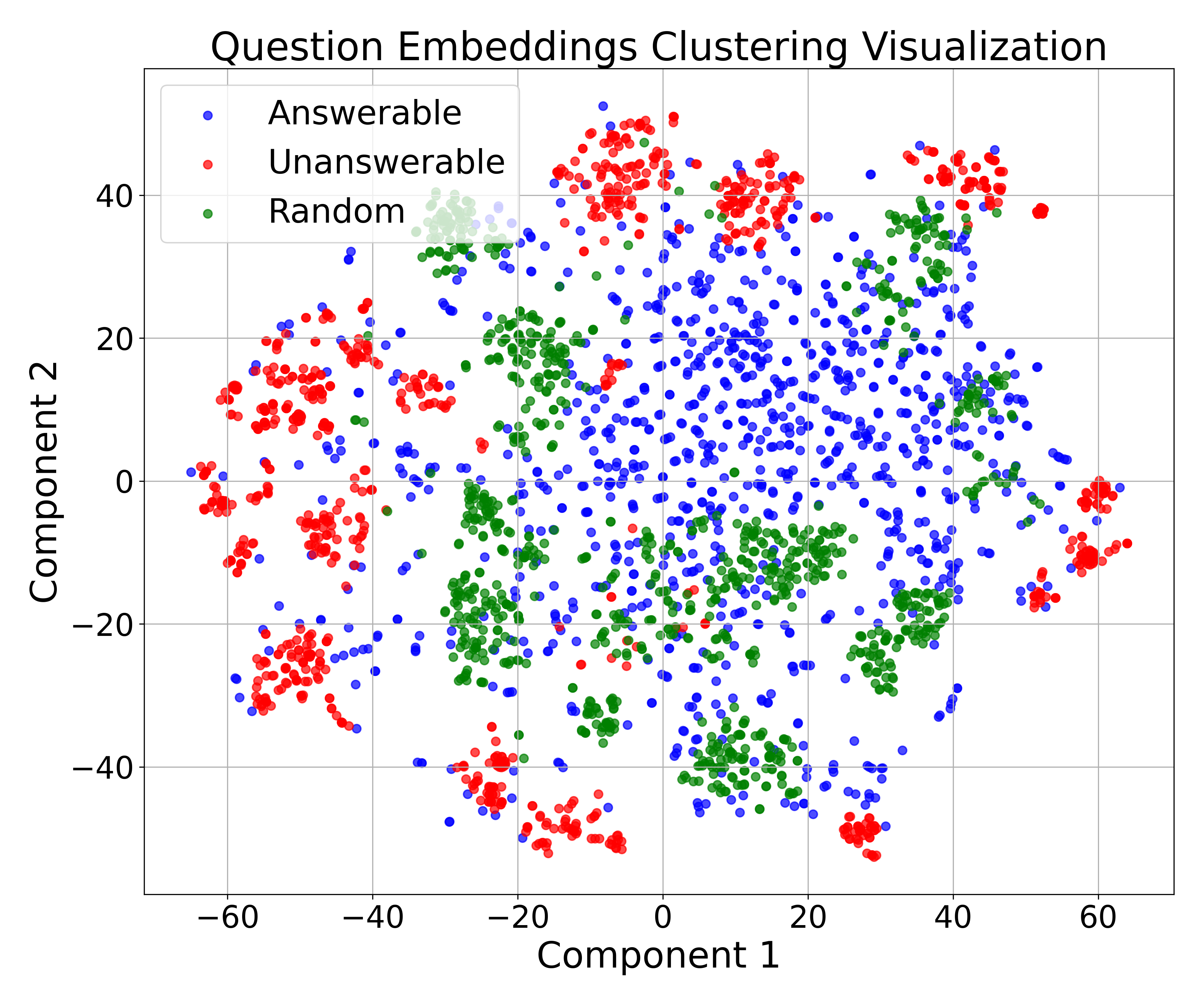}
    \caption{t-SNE visualization of question embeddings. Blue: answerable questions (within-knowledge boundary) form a central cluster. Red: unanswerable questions (beyond-knowledge boundary) form a surrounding band. Green: random questions are diffusely scattered. This structure supports the non-triviality of KBD-generated samples.}
    \label{fig:tsne}
\end{figure}

Then to verify that the questions generated by our KBD algorithm are non-trivial, that is, they are not random or easily separable but instead lie meaningfully near the model’s decision boundary, we analyze their distribution in the semantic embedding space. Specifically, we embed 1,000 questions from each of three categories: \textit{answerable} (within-knowledge boundary), \textit{unanswerable} (beyond-knowledge boundary), and \textit{randomly generated}. We use a sentence-level embedding model to obtain high-dimensional representations, which we then cluster before applying t-SNE for 2D visualization. As shown in Figure~\ref{fig:tsne}, the structure supports the validity of our identified knowledge boundary: The questions form distinct clusters that meaningfully distributed near the knowledge boundary, demonstrating that KBD-generated samples are informative and non-trivial.

\subsection{KBD-Generated Examples and Boundary Analysis}

Using KBD, we automatically generate questions that the target LLM can or cannot answer with high confidence.  Table~\ref{table:KBD_question_set_structure} shows typical \textit{within-knowledge} and \textit{beyond-knowledge} items, demonstrating that our algorithm accurately identifies the knowledge frontier.

We applied the method to both \emph{specialized technical domains}—such as clinical medicine, biotechnology \cite{singhal2022largelanguagemodelsencode}, and fundamental science concepts \cite{wang2024scibenchevaluatingcollegelevelscientific}—and four broad academic areas: social sciences, natural sciences, applied sciences and humanities.  In every setting, KBD located coherent boundaries; further examples are provided in Appendix B.

The questions whose entropies lie in the narrow transition band (\(40<E<170\)) reveal the upper and lower bounds of the model’s knowledge, where the answers become ambiguous. Some examples are provided in Appendix C.

\subsection{Comparison of KBD-generated Dataset with Other Human-generated Dataset}

\begin{table}[t]
\renewcommand{\arraystretch}{1.5}
\centering
\begin{tabular}{@{}lcccc@{}}
\toprule
\textbf{Model}       & \multicolumn{2}{c}{\textbf{KBD}(Ours)} & \multicolumn{2}{c}{\textbf{KUQ}} \\ \cmidrule(lr){2-3} \cmidrule(lr){4-5}
                     & \textbf{EER} & \textbf{F1}      & \textbf{EER} & \textbf{F1}      \\ \midrule
LlaMA 7B-Chat             & 0.239        & 0.725            & 0.301        & 0.732            \\
ChatGLM3-6B          & 0.267        & 0.507            & 0.325        & 0.484            \\
ChatGLM2-6B          & 0.508        & 0.431            & 0.514        & 0.449            \\ \bottomrule
\end{tabular}
\caption{Performance comparison of different LLMs on KBD-generated dataset and KUQ dataset. Metrics include Equal Error Rate (EER) and F1 score (lower EER, better performance, higher F1 scores, better performance). The table demonstrates that the EER and F1 scores of our KBD-generated dataset show similar performance to the KUQ dataset across different models, validating that our KBD-generated dataset comparable to human-generated datasets.}
\label{table:model_comparison}
\end{table}

Our KBD-generated dataset achieves performance comparable to that of other human-generated datasets. To validate this, we compared our question set with the KUQ dataset \cite{amayuelas2024knowledgeknowledgeexploringknownunknowns}. Following KUQ, they define a similarity function, \( f_{\text{sim}} \), as a binary metric between the generated text (\( t_i \)) and some reference text (\( \text{ref}_i \)) to be 1 if they express the same content or 0 if they do not (the reference texts are a predefined set of phrases that encompass general uncertainty, and the full list can be found in Appendix D). Specifically, if the reference text is contained in the generated text or the similarity measured with SimCSE \shortcite{gao2022simcsesimplecontrastivelearning} exceeds a threshold \( \tau \), the function returns 1:

\begin{equation}
\text{Sim}_i = f_{\text{sim}}(\text{t}_i, \text{ref}_i)
\end{equation}

In our evaluation, two metrics are adopted: the F1 score and the Equal Error Rate (EER). The F1 score, derived from the similarity metric, evaluates the positive class (e.g., unknown questions or the chosen category). The EER measures the balance between false acceptance and false rejection rates, providing a holistic view of the LLM's performance on the dataset.

As shown in Table~\ref{table:model_comparison}, models exhibit similar performance in the KBD-generated and KUQ datasets. This validates the effectiveness of our automated question generation framework. Additionally, consistent with previous findings, models with larger parameter counts(e.g., LlaMA 7B-Chat) tend to perform better in boundary recognition tasks.

\subsection{Comparison of KBD Algorithm with Human Expert Questioning and Random Questioning}

We compare our KBD algorithm with two baselines: expert questioning and random questioning, evaluated using the entropy metric. Expert questioning represents the upper bound, while random questioning serves as the lower bound.

For expert questioning, domain experts iteratively asked questions to explore the LLM’s knowledge boundary. For random questioning, questions were randomly selected from the science book \emph{Hundred Thousand Whys} and posed without considering context or relevance.

\begin{figure}[t] 
    \centering
    \includegraphics[width=\linewidth]{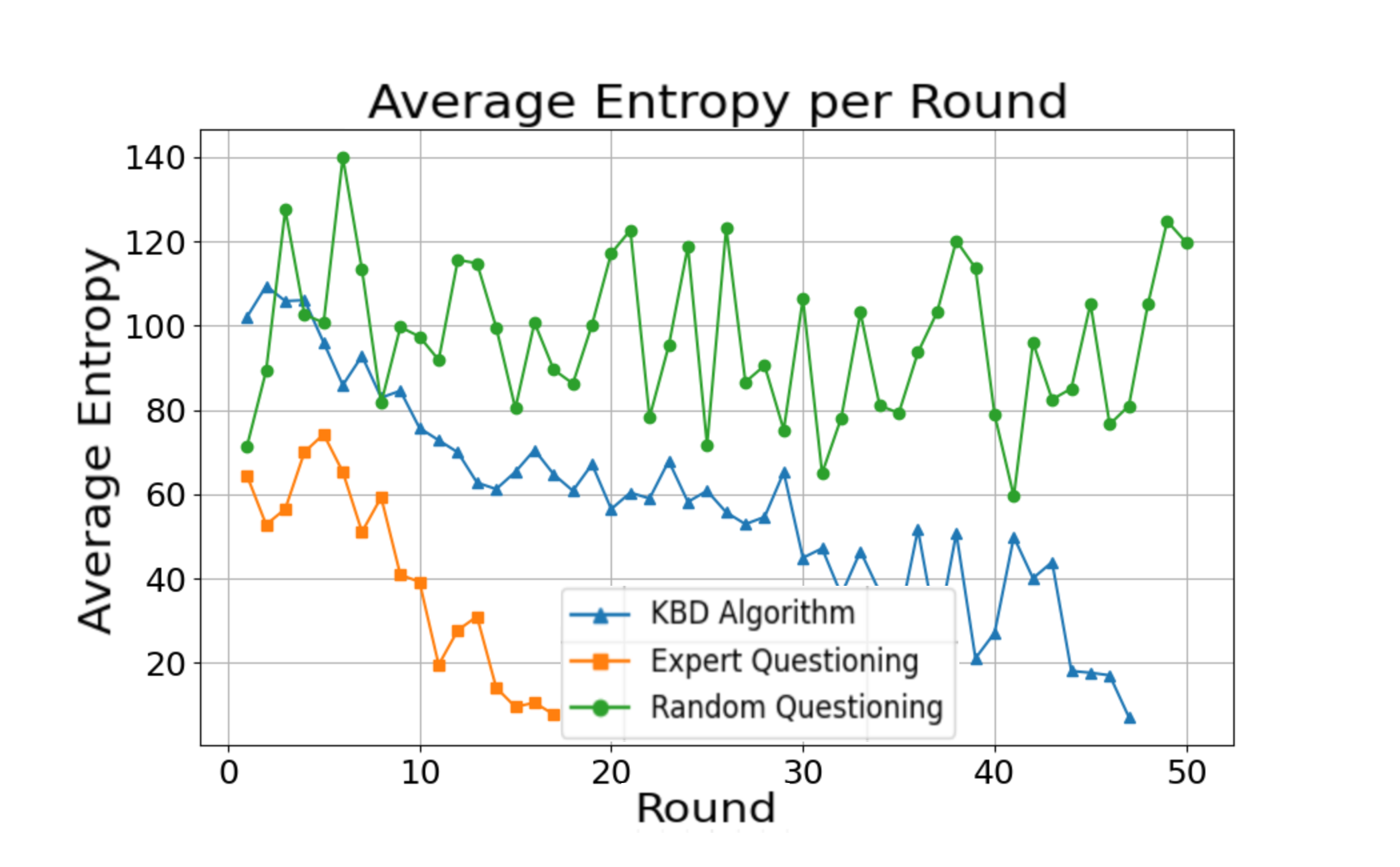} 
    \caption{The figure illustrates the average entropy over 50 rounds across 1000 episodes for our KBD algorithm, the expert questioning baseline, and the random questioning baseline. It shows that both KBD algorithm and expert questioning effectively explore the LLM’s knowledge boundary. However, the random questioning baseline fails to converge or reach the knowledge boundary.} 
    \label{fig:example2} 
\end{figure}

As shown in Figure~\ref{fig:example2}, both our proposed KBD algorithm and the expert questioning baseline exhibit a decreasing trend in entropy over successive rounds of interaction. They successfully identify the within-knowledge boundary of the target LLM (entropy \( \leq 40 \)) and maintain convergence in subsequent responses. In contrast, random questioning fails to reach the boundary within 50 rounds and exhibits significant entropy fluctuations.

This comparison demonstrates the effectiveness of the KBD in identifying knowledge boundaries. Although it is slower than expert questioning, it significantly outperforms random questioning, providing a scalable and automated solution for exploring LLM knowledge boundaries.

\subsection{Convergence of KBD Algorithm}

\begin{figure}[t] 
    \centering
    \includegraphics[width=\linewidth]{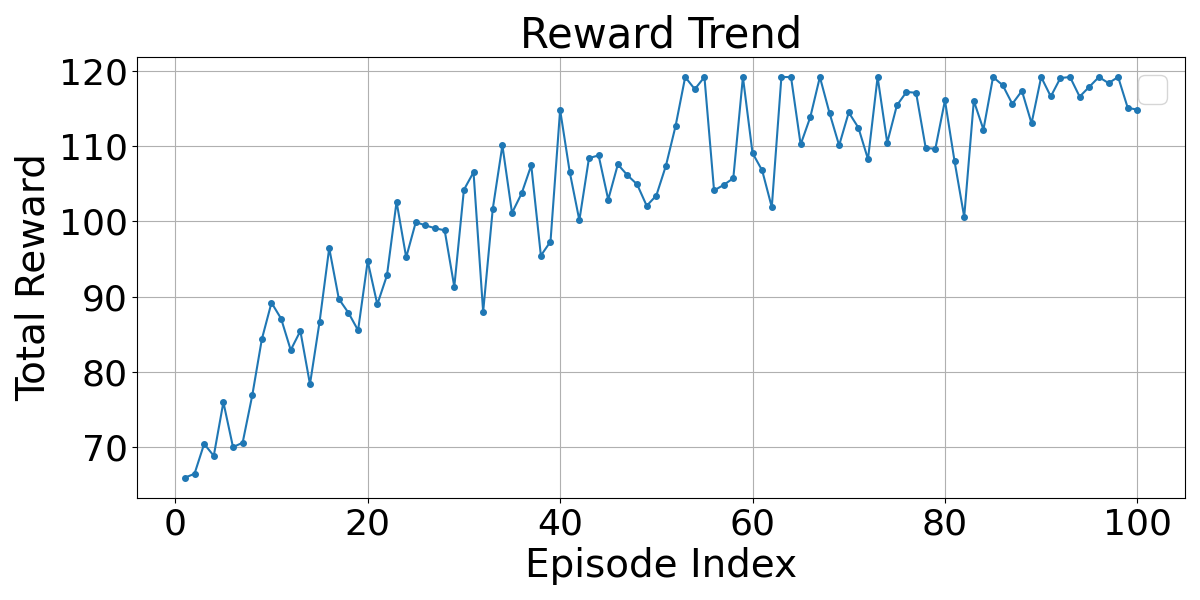} 
    \caption{The cumulative reward per episode increases with training episodes and eventually converges. This indicates that KBD is consistently learning to optimize strategies.} 
    \label{fig:reward_convergence} 
\end{figure}

This section highlights the effectiveness of our algorithm in learning an optimal strategy during training. As shown in Figure~\ref{fig:reward_convergence}, the cumulative reward stabilizes between 110 and 120 after approximately 50 episodes, indicating convergence. This shows that the algorithm progressively optimizes its strategy to achieve higher rewards and effectively explore the knowledge boundaries of the LLM.

\section{Conclusion}

In this work, we introduce a practical and operational definition of the knowledge boundary for LLMs, distinguishing between the questions that the model can confidently answer (\textit{within-knowledge boundary}) and those it cannot (\textit{beyond-knowledge boundary}). Building upon this foundation, we propose \textbf{Knowledge Boundary Discovery (KBD)}, the first reinforcement learning framework that dynamically explores and uncovers these boundaries through entropy-guided interaction. Through extensive experiments, we demonstrate the effectiveness and superiority of entropy as a confidence estimation metric, outperforming traditional probability-based baselines in identifying unanswerable questions and capturing model uncertainty. Furthermore, the KBD-generated dataset that we construct is both well structured and meaningful, achieving performance comparable to human-generated datasets. These findings validate the reliability of our proposed framework and the soundness of our knowledge boundary definition for LLMs.


\section*{Acknowledgments}
This work is supported by the Science Challenge Project (Grant No.~TZ2025008), the Science Foundation of China University of Petroleum, Beijing (Grant No.~2462023YJRC024) and the Frontier
Interdisciplinary Exploration Research Program of China University of
Petroleum, Beijing (Grant No.~2462024XKQY003). Zhongqi Lu is the
corresponding author.


\bibliography{main}

\end{document}